\documentclass[11pt,a4paper,logo]{lumia}

\usepackage[round,authoryear]{natbib}
\usepackage{multirow}
\usepackage{array}
\definecolor{bigaired}{RGB}{156,0,0}
\definecolor{color5}{HTML}{006795}
\hypersetup{
    colorlinks=true,
    linkcolor=bigaired, 
    citecolor=color5,   
    urlcolor=color5     
}
\newsavebox{\fusioncaptionbox}
\newsavebox{\locationcaptionbox}
\newlength{\sharedcaptionheight}

\setheadertext{LUMIA Lab}
\correspondingemail{\emailicon~hyq718@sjtu.edu.cn, boyizeng@sjtu.edu.cn, lin.zhouhan@gmail.com \quad $^*$ Equal Contribution \quad $^\ddagger$ Corresponding Author.}
\githublink{https://github.com/LUMIA-Group/Dual-Slot}

\title{Dual-Slot: From Serial Loops to Parallel Slots in Language Models}
\author{%
Yiqin Hao$^{1,2*}$, Zitong Wang$^{1,2,4*}$, Boyi Zeng$^{1}$, He Li$^{1}$,
Shixiang Song$^{1}$, Feichen Song$^{1}$, Yifan Liu$^{1}$, Ran Tao$^{2}$, Bryan Dai$^{2}$, Zhouhan Lin$^{1,3\ddagger}$\\
$^{1}$LUMIA Lab, School of Artificial Intelligence, Shanghai Jiao Tong University \quad$^{2}$IQuest Research \quad $^3$ Shanghai AI Laboratory \quad$^{4}$Sun Yat-sen University
}
\begin{document}

\begin{abstract}
Looped language models increase computational depth by repeatedly applying
shared layers, offering a way to improve model quality without adding
parameters. However, sequential dependencies between loop iterations
increase autoregressive decoding latency. We introduce \textbf{Dual-Slot},
an architecture that reorganizes serial looped computation into parallel
latent and prediction slots. Rather than waiting for the current token's
latent computation to finish, the prediction slot combines the previous
token's deep latent state with the current representation, while the latent
slot produces a state for the next decoding step. This shifts the latent
dependency across adjacent tokens, replacing two sequential module
invocations with a single paired invocation while preserving standard
autoregressive generation. To preserve token-level parallelism during
training, we use a small number of Jacobi iterations to refine latent states
across the sequence in parallel. Experiments on Qwen-style dense and MoE
models show consistent improvements in language modeling and downstream
performance. On the 1B dense model, Dual-Slot reduces Pile validation
perplexity from 7.25 to 7.00 and improves average five-shot accuracy across
nine task configurations by 2.26 percentage points. After supervised
fine-tuning, it improves zero-shot GSM8K accuracy by 5.83 and 3.49
percentage points for the 1B dense and A0.6B MoE models, respectively,
while maintaining near-vanilla decoding latency.
\end{abstract}
\maketitle

\section{Introduction}
\label{sec:introduction}

Increasing the computation applied to each token offers a way to improve
language model quality without increasing model size. Looped language
models implement this idea by repeatedly applying shared layers, increasing
effective computational depth through parameter
reuse~\citep{dehghani2018universal,giannou2023looped,geiping2026scaling}.

However, parameter sharing does not eliminate the cost of repeated
computation: successive loop iterations remain sequentially dependent
and can make autoregressive decoding several times slower than a vanilla
Transformer with the same parameter count. The challenge is therefore
to benefit from recurrent computation without paying for multiple serial
loop invocations at every decoding step.

To address this bottleneck, we introduce \textbf{Dual-Slot}, an architecture
that reorganizes serial looped computation into parallel latent and
prediction slots. Rather than waiting for the current token's latent
computation to finish, the prediction slot combines the previous token's
already-computed latent state with the current representation. Meanwhile,
the latent slot processes the current representation to produce a state
for the next decoding step. This shifts the latent dependency across
adjacent tokens, replacing two sequential module invocations with a
single paired invocation while preserving standard autoregressive
generation. 

While this design is natural during autoregressive decoding, it introduces
a sequential dependency during teacher-forced training, where the latent
states of preceding tokens are not yet available. We address this challenge
using a small number of Jacobi iterations that refine latent states at all
positions in parallel. This preserves token-level training parallelism
while approximating the recurrent states used during decoding. We implement
Dual-Slot in the middle third of a Transformer, leaving the prefix and
suffix as single-pass components.

We train and evaluate a series of Qwen-style dense and MoE models \citep{yang2025qwen3}.
Dual-Slot consistently improves language modeling and average downstream
accuracy over the corresponding vanilla models. For the 1B dense model,
Pile validation perplexity decreases from 7.25 to 7.00, while average
five-shot accuracy across nine task configurations improves by 2.26
percentage points. After supervised fine-tuning, Dual-Slot improves
zero-shot GSM8K accuracy by 5.83 and 3.49 percentage points for the
1B dense and A0.6B MoE models, respectively. These improvements are
achieved while maintaining near-vanilla decoding latency.

\begin{figure}[t]
    \centering
    \includegraphics[width=\linewidth]{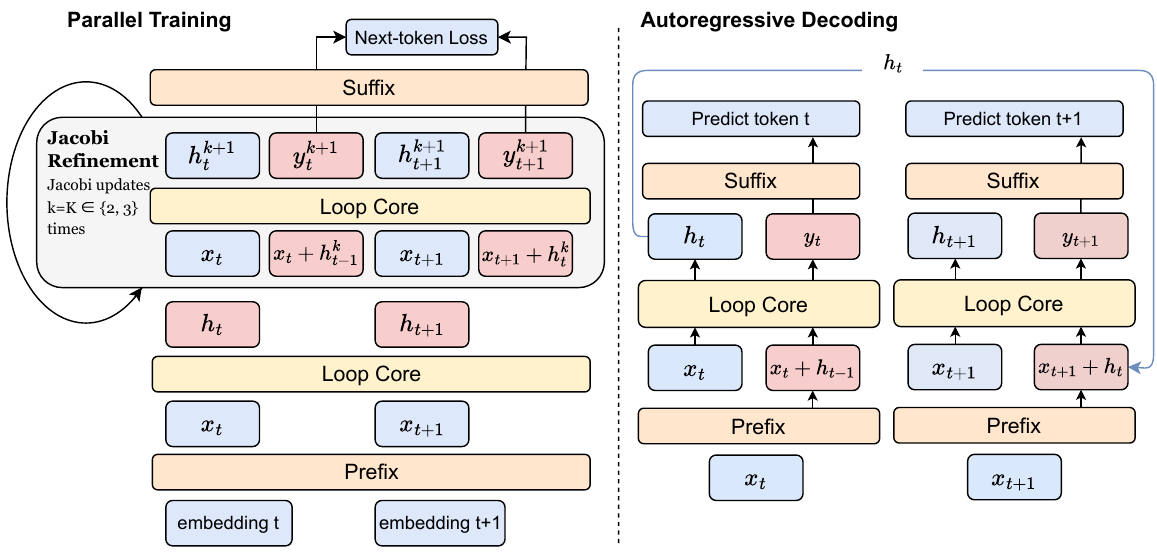}
\caption{\textbf{Dual-Slot training and inference.}
1) \textbf{Parallel Training (left).} The prefix processes the full
sequence once, and an initial warm pass through the core produces the initial latent slots. Jacobi updates then refine the slots at all positions in parallel, with next-token loss computed only from the final prediction states.
2) \textbf{Autoregressive Decoding (right).} At decoding step $t$, the shared Loop Core processes $x_t$ and $x_t+h_{t-1}$ in one invocation. The prediction-slot output $y_t$ predicts token $t+1$, while the latent-slot output $h_t$ is carried to the next step.}
\label{fig:dualslot}
\end{figure}

\section{Related Work}
\paragraph{Looped language models.} Parameter sharing allows a Transformer to reuse the same layers, increasing computation depth without adding parameters. Early work applies shared layers recurrently across depth \citep{dehghani2018universal}. Later studies use repeated shared modules to improve language modeling and reasoning \citep{giannou2023looped,yang2024looped}, while recurrent-depth models increase computation within the latent space \citep{geiping2026scaling}. Some methods reuse intermediate latent representations by feeding continuous outputs back into the model \citep{zeng2026ponderlm} or refining the current hidden state as a latent thought \citep{zeng2025ponderlm}. However, these computations remain sequential, so decoding latency grows with the number of loops.

\paragraph{Efficient looped inference.} Several methods reduce this cost through adaptive computation. Some determine whether to continue from the current state \citep{banino2021pondernet}, while token-level methods allocate different amounts of computation to different tokens \citep{li2026ponderlm}. Dynamic routing and separate KV caches can support variable loop depths \citep{yu2026t}, but they introduce irregular execution and complicate batching, scheduling, and cache management. Other methods stagger computation across adjacent decoding steps \citep{cutler2025stagformer}, execute loop stages from different tokens in one model invocation \citep{wu2025parallel}, or use a parameter-shared parallel prediction stream \citep{langford2026free}. However, their information flow remains tied to a predefined stage schedule. During autoregressive decoding, the previous token's latent state is available before the current step~\citep{fan2020addressing,tang2026turbo,cai2026t,huang2026latent} and can support prediction while the current latent state is computed. Dual-Slot uses this property to compute a latent slot and a prediction slot in one shared-module invocation, with Jacobi iteration preserving token-level parallelism during training.

\begin{figure}[!t]
    \centering
    \includegraphics[width=\linewidth]{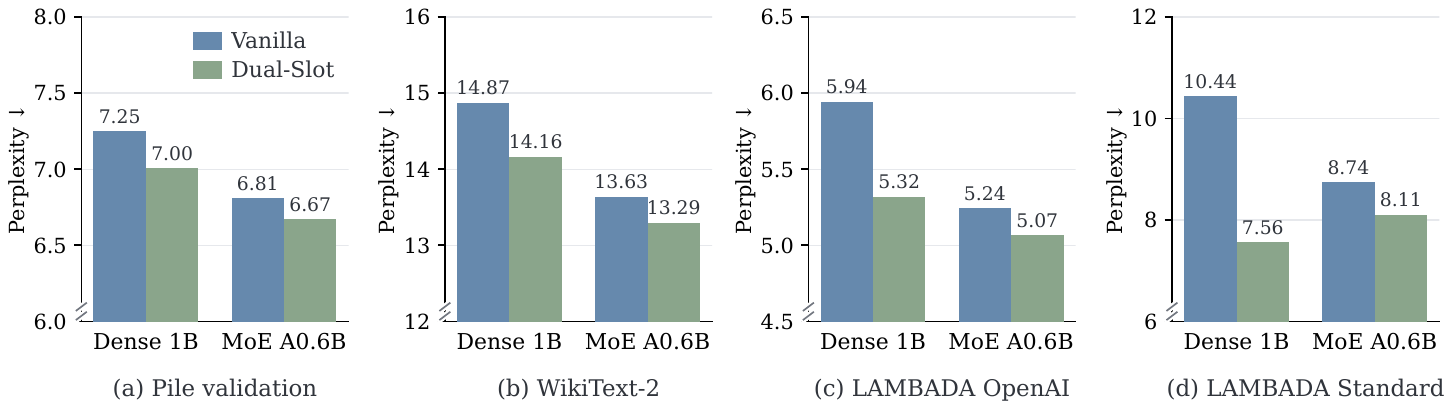}
    \caption{\textbf{Language modeling performance of the 1B dense and
    A0.6B MoE models.}
    From left to right: perplexity on Pile validation, WikiText-2
    (word-level), LAMBADA-OpenAI, and LAMBADA-Standard.
    Lower is better. Dual-Slot improves over Vanilla across all four
    benchmarks for both model families.}
    \label{fig:language-modeling-ppl}
\end{figure}

\section{Method}

Looped language models allocate additional computation to each token by
repeatedly applying a shared module. Consider a two-loop computation and let
$F$ denote the looped module. Omitting the causal context for notational
simplicity, the computation can be written schematically as:
\begin{equation}
    \text{Loop 1:}\quad h_t = F(x_t);
    \qquad
    \text{Loop 2:}\quad y_t = F(h_t).
    \label{eq:serial-loop}
\end{equation}
Here, $x_t$ is the representation of token $t$ entering the looped module,
$h_t$ is an intermediate latent representation, and $y_t$ is used to predict
token $t+1$. In this conventional formulation, Loop 2 depends on the output
of Loop 1 and therefore cannot begin until Loop 1 has finished. More
generally, $L$ loops require $L$ sequential applications of $F$ along the
autoregressive decoding path.

To address the decoding latency introduced by serial looping, we propose
Dual-Slot, which introduces a latent slot and a prediction slot for each
token. Neither slot requires the other's final output at the current step,
allowing the two slots to advance through the shared module in parallel.

We first describe this decoding procedure and its parallel training rule while
treating $F$ as an abstract looped module. We then instantiate $F$ as a
partial loop over the middle layers of a Transformer in~\autoref{sec:partial-loop}.

\subsection{Dual-Slot Parallel Decoding}
\label{sec:dual-slot-decoding}

At decoding step $t$, we define the two slot inputs as:
\begin{equation}
    \text{Latent slot:}\quad x_t;
    \qquad
    \text{Prediction slot:}\quad
    \widetilde{x}_t
    = \operatorname{Mix}(x_t, h_{t-1})
    = x_t + h_{t-1},
    \label{eq:history-mixing}
\end{equation}
where $x_t$ is the current representation and $h_{t-1}$ is the preceding
token's latent-slot output. The carried history is set to zero at sequence
boundaries.

The two slots share token $t$'s position ID and are ordered causally, with
the latent slot first. Given the attention key--value cache
$\mathrm{KV}_{<t}$ from preceding steps, we evaluate them together:
\begin{equation}
    (h_t, y_t)
    = F\!\left([x_t, \widetilde{x}_t];\, \mathrm{KV}_{<t}\right).
    \label{eq:dual-slot-decoding}
\end{equation}
Here, $h_t$ is retained for the next decoding step, while $y_t$ is used to
predict token $t+1$. During autoregressive decoding, $h_{t-1}$ is already
available from the preceding step, so both slot inputs are known before
invoking $F$ and the two slots can be processed in parallel.

\begin{figure}[!t]
    \centering
    \includegraphics[width=\linewidth]{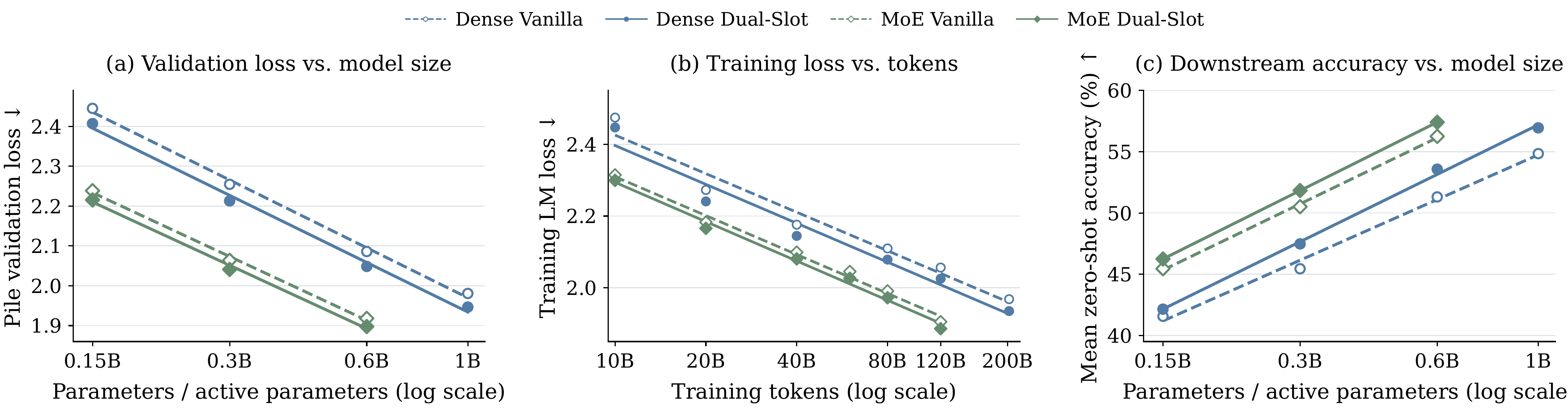}
    \caption{\textbf{Scaling behavior of Dual-Slot.}
    (a) Pile validation loss versus model size.
    (b) Loss versus training tokens for the 1B dense and A0.6B MoE models.
    (c) Average zero-shot accuracy over nine downstream task
    configurations versus model size.
    Model size denotes the nominal dense parameter count or the
    MoE active parameter count.
    Lower is better in (a) and (b); higher is better in (c).}
    \label{fig:scaling}
\end{figure}

\subsection{Parallel Training via Jacobi Iteration}
\label{sec:jacobi-training}

Sequential decoding naturally provides $h_{t-1}$ before token $t$ is
processed. Teacher-forced pretraining, however, processes all tokens in a
sequence simultaneously, before their latent-slot outputs are available.
Naively reproducing the decoding procedure would create a left-to-right chain
$h_1 \rightarrow h_2 \rightarrow \cdots \rightarrow h_T$ and remove the
token-level parallelism of Transformer training. To preserve token-level
parallelism while approximating the latent states induced by sequential
inference, we use Jacobi iteration to refine all positions in parallel for a
small number of rounds~\citep{saad2003iterative}.

\paragraph{Initial state estimation.}
Given the unexpanded input sequence $X = [x_1, \ldots, x_T]$, we first apply
the looped module once without history injection:
\begin{equation}
    H^{(0)}
    =
    [h_1^{(0)}, \ldots, h_T^{(0)}]
    =
    F(X).
    \label{eq:warm-pass}
\end{equation}
Using $H^{(0)}$, we construct the initial prediction-slot inputs in parallel
and insert them after their corresponding latent-slot inputs:
\begin{equation}
    \widetilde{x}_t^{(0)}
    =
    x_t + h_{t-1}^{(0)},
    \qquad
    \widetilde{X}^{(0)}
    =
    [x_1, \widetilde{x}_1^{(0)}, \ldots,
     x_T, \widetilde{x}_T^{(0)}].
    \label{eq:jacobi-mixing}
\end{equation}
At sequence boundaries, the shifted history state is set to zero. We refer to~\autoref{eq:warm-pass} as the \emph{warm pass}; it provides the initial
estimates used to construct $\widetilde{X}^{(0)}$ and is not counted as a
Jacobi update.

\paragraph{Parallel state updates.}
At Jacobi round $k$, beginning with $k=0$, we process the current interleaved
sequence in one application of $F$:
\begin{equation}
    \bigl(H^{(k+1)}, Y^{(k+1)}\bigr)
    =
    F\!\left(\widetilde{X}^{(k)}\right),
    \label{eq:jacobi-update}
\end{equation}
where $H^{(k+1)}$ and $Y^{(k+1)}$ collect the outputs at the latent and
prediction slots, respectively. We then use the new latent-slot outputs to
update all prediction-slot inputs in parallel and construct the sequence for
the next round:
\begin{equation}
    \widetilde{x}_t^{(k+1)}
    =
    x_t + h_{t-1}^{(k+1)},
    \qquad
    \widetilde{X}^{(k+1)}
    =
    [x_1, \widetilde{x}_1^{(k+1)}, \ldots,
     x_T, \widetilde{x}_T^{(k+1)}].
    \label{eq:jacobi-reconstruction}
\end{equation}

\paragraph{Training objective.}
After $K$ updates, we apply the standard next-token cross-entropy objective
only to the final prediction states $Y^{(K)}$:
\begin{equation}
    \mathcal{L}_{\mathrm{LM}}
    =
    -\sum_{t=1}^{T-1}
    \log p_{\theta}\!\left(
        \text{token}_{t+1}
        \mid
        y_t^{(K)}
    \right).
    \label{eq:lm-objective}
\end{equation}
The latent-slot outputs receive no auxiliary language-modeling loss. To expose
the model to variable refinement depths, we synchronously sample
$K$ from $\{2,3\}$ for each training batch \citep{zeng2025ponderlm}. We report the empirical
convergence behavior in~\autoref{sec:Experiments}.

\subsection{Partial-Loop Implementation}
\label{sec:partial-loop}

In practice, to balance model performance and computational cost, we apply
Dual-Slot in a partial-loop form rather than looping the entire model. Only a
contiguous set of middle Transformer layers is looped. The complete model can
therefore be viewed as three components: a Prefix, a Loop Core, and a Suffix.

For each token, the Prefix is evaluated once to produce the loop input:
\begin{equation}
    x_t
    =
    \operatorname{Prefix}\!\left(
        \operatorname{Embed}(\text{token}_t)
    \right).
    \label{eq:prefix}
\end{equation}
The Loop Core serves as $F$ in~\autoref{eq:dual-slot-decoding}--\autoref{eq:jacobi-update}: only these middle
layers operate on the paired latent and prediction slots. We then discard the
latent stream and pass only the prediction output through the Suffix and the
language modeling head:
\begin{equation}
    \operatorname{logits}_t
    =
    \operatorname{LMHead}\!\left(
        \operatorname{Suffix}(y_t)
    \right).
    \label{eq:suffix}
\end{equation}
Thus, token encoding in the Prefix and prediction decoding in the Suffix are
performed once, while recurrent latent computation is confined to an
intermediate representation space.

\begin{table}[!t]
    \centering
    \caption{\textbf{Downstream performance of the 1B dense and A0.6B MoE
    models.} Zero-shot and five-shot accuracy (\%).
    Avg. is the unweighted mean over the nine task configurations.
    Bold indicates the best score within each model and shot setting,
    including ties.}
    \label{tab:downstream-main}

    \footnotesize
    \setlength{\tabcolsep}{2.5pt}
    \renewcommand{\arraystretch}{1.18}

    \begin{tabularx}{\linewidth}{
        @{}
        c
        l
        *{10}{>{\centering\arraybackslash}X}
        @{}
    }
        \toprule

        \multirow{2}{*}{Model}
        & \multirow{2}{*}{Method}
        & \multicolumn{2}{c}{LAMBADA}
        & \multirow{2}{*}{PIQA}
        & \multirow{2}{*}{\shortstack{Wino\\Grande}}
        & \multirow{2}{*}{SciQ}
        & \multirow{2}{*}{\shortstack{Hella\\Swag}}
        & \multicolumn{2}{c}{ARC}
        & \multirow{2}{*}{\shortstack{RACE\\high}}
        & \multirow{2}{*}{Avg.} \\

        \cmidrule(lr){3-4}
        \cmidrule(lr){9-10}

        &
        & OpenAI
        & Standard
        &
        &
        &
        &
        & Easy
        & Challenge
        &
        & \\

        \midrule

        \multicolumn{12}{c}{\textbf{Zero-shot}} \\
        \midrule

        \multirow{2}{*}{\shortstack{1B\\Dense}}
& Vanilla
& 62.64
& 52.42
& 71.93
& \textbf{60.38}
& 89.30
& 42.09
& 60.82
& \textbf{27.56}
& 35.31
& 55.83 \\

& Dual-Slot
& \textbf{64.95}
& \textbf{56.69}
& \textbf{72.58}
& 59.75
& \textbf{90.20}
& \textbf{43.74}
& \textbf{61.95}
& 27.47
& \textbf{36.56}
& \textbf{57.10} \\

        \addlinespace[4pt]

        \multirow{2}{*}{\shortstack{A0.6B\\MoE}}
        & Vanilla
        & 64.43
        & 52.61
        & 72.85
        & \textbf{59.04}
        & 88.60
        & 44.12
        & 63.05
        & 26.79
        & \textbf{34.83}
        & 56.26 \\

        & Dual-Slot
        & \textbf{64.70}
        & \textbf{55.62}
        & \textbf{73.23}
        & 58.33
        & \textbf{89.00}
        & \textbf{45.33}
        & \textbf{65.07}
        & \textbf{31.06}
        & 34.35
        & \textbf{57.41} \\

        \midrule

        \multicolumn{12}{c}{\textbf{Five-shot}} \\
        \midrule

        \multirow{2}{*}{\shortstack{1B\\Dense}}
        & Vanilla
        & 57.62
        & 49.18
        & 72.58
        & 60.06
        & \textbf{93.40}
        & 41.68
        & 63.38
        & 29.10
        & 34.55
        & 55.73 \\

        & Dual-Slot
        & \textbf{58.86}
        & \textbf{56.14}
        & \textbf{73.23}
        & \textbf{62.59}
        & \textbf{93.40}
        & \textbf{43.75}
        & \textbf{65.70}
        & \textbf{30.97}
        & \textbf{37.22}
        & \textbf{57.98} \\

        \addlinespace[4pt]

        \multirow{2}{*}{\shortstack{A0.6B\\MoE}}
        & Vanilla
        & 57.97
        & 49.47
        & 73.78
        & 60.22
        & 92.20
        & 43.88
        & 65.78
        & 30.46
        & \textbf{35.02}
        & 56.53 \\

        & Dual-Slot
        & \textbf{58.78}
        & \textbf{52.30}
        & \textbf{74.10}
        & \textbf{61.09}
        & \textbf{94.60}
        & \textbf{45.65}
        & \textbf{67.97}
        & \textbf{32.94}
        & 34.64
        & \textbf{58.01} \\

        \bottomrule
    \end{tabularx}
\end{table}

\section{Experiments}
\label{sec:Experiments}

We evaluate Dual-Slot from three perspectives: language modeling and
downstream performance, scaling across model sizes and training budgets,
and the convergence of its Jacobi refinement procedure. Our main results
focus on a qwen-style 1B dense model and an A0.6B mixture-of-experts (MoE) model. At each scale, we compare
Dual-Slot with the corresponding vanilla Transformer baseline.


\subsection{Main Results}
\label{sec:main-results}

\subsubsection{Language Modeling Ability}
\label{sec:language-modeling}

We evaluate the 1B dense and A0.6B MoE models on Pile validation~\citep{gao2020pile},
WikiText-2~\citep{merity2016pointer}, LAMBADA-OpenAI, and
LAMBADA-Standard~\citep{paperno2016lambada}, using word-level perplexity
for WikiText-2. As shown in~\autoref{fig:language-modeling-ppl}, Dual-Slot
consistently improves language modeling performance over Vanilla across
all four benchmarks, achieving lower perplexity for both dense and MoE models.

\subsubsection{Downstream Performance}
\label{sec:downstream-main}

We evaluate the 1B dense and A0.6B MoE models on nine downstream task
configurations: LAMBADA-OpenAI, LAMBADA-Standard~\citep{paperno2016lambada}, PIQA~\citep{bisk2020piqa}, WinoGrande~\citep{sakaguchi2021winogrande}, SciQ~\citep{welbl2017crowdsourcing},
HellaSwag~\citep{zellers2019hellaswag}, ARC-Easy, ARC-Challenge~\citep{clark2018think}, and RACE-high~\citep{lai2017race}.~\autoref{tab:downstream-main} reports zero-shot and five-shot accuracy,
together with the unweighted average over all nine configurations.

In zero-shot evaluation, Dual-Slot improves average accuracy from
55.83\% to 57.10\% for the 1B dense model and from 56.26\% to 57.41\%
for the A0.6B MoE model, corresponding to gains of 1.27 and 1.15
percentage points, respectively. The gains also hold in the five-shot
setting, where average accuracy increases 2.26 and 1.48
percentage points. Although individual task gains vary, Dual-Slot
consistently improves aggregate downstream performance across both
model families and shot settings.

\subsubsection{Mathematical Reasoning after Fine-Tuning}
\label{sec:gsm8k}

We evaluate mathematical reasoning on GSM8K~\citep{cobbe2021training} after supervised fine-tuning
of the pretrained models.~\autoref{tab:gsm8k} reports final-answer
exact-match accuracy under both zero-shot and eight-shot evaluation.

Dual-Slot consistently improves GSM8K performance across both model
families and prompting settings. For the 1B dense model, zero-shot and
eight-shot accuracy improve by 5.83 and 4.10 percentage points,
respectively. The A0.6B MoE model shows similar gains of 3.49 and
3.87 points. These results indicate that the improvements from
Dual-Slot persist after supervised fine-tuning and extend to
mathematical reasoning.

\begin{table}[!t]
    \centering
    \caption{\textbf{Mathematical reasoning after supervised fine-tuning.}
    GSM8K test final-answer EM (\%) under zero-shot and eight-shot
    prompting for the 1B dense and A0.6B MoE models.
    Dual-Slot consistently improves accuracy across both model families
    and evaluation settings.}
    \label{tab:gsm8k}

    \small
    \setlength{\tabcolsep}{7pt}
    \renewcommand{\arraystretch}{1.12}
    \begin{tabular}{@{}clcccc@{}}
        \toprule
        \multirow{2}{*}{Model}
        & \multirow{2}{*}{Method}
        & \multicolumn{2}{c}{Zero-shot}
        & \multicolumn{2}{c}{Eight-shot} \\
        \cmidrule(lr){3-4}
        \cmidrule(lr){5-6}
        & & EM (\%) $\uparrow$ & Gain (pp.)
          & EM (\%) $\uparrow$ & Gain (pp.) \\
        \midrule

        1B Dense & Vanilla
        & 13.12 & --
        & 10.84 & -- \\

        1B Dense & Dual-Slot
        & \textbf{18.95} & $+5.83$
        & \textbf{14.94} & $+4.10$ \\

        \midrule

        A0.6B MoE & Vanilla
        & 11.60 & --
        & 7.96 & -- \\

        A0.6B MoE & Dual-Slot
        & \textbf{15.09} & $+3.49$
        & \textbf{11.83} & $+3.87$ \\

        \bottomrule
    \end{tabular}
\end{table}

\subsection{Scaling Analysis}
\label{sec:scaling}

We study the scaling behavior of Dual-Slot using dense models with
150M, 300M, 600M, and 1B parameters, and MoE models with A0.15B, A0.3B,
and A0.6B active parameters, with tpp200 (token-per-parameter).~\autoref{fig:scaling} summarizes
validation loss across model sizes, loss curves over training tokens,
and average downstream accuracy across model sizes.

\paragraph{Scaling with model size.}
\autoref{fig:scaling}(a) reports final Pile validation loss.
Dual-Slot outperforms Vanilla at every evaluated scale,
reducing loss by about 0.035 for dense models and 0.020 for MoE models. These improvements persist as model size increases.

\paragraph{Scaling with training tokens.}
~\autoref{fig:scaling}(b) plots loss against training tokens for the
1B dense and A0.6B MoE models. Dual-Slot maintains lower loss at every
reported checkpoint for both model families, with the advantage
persisting through approximately 202B and 120B training tokens.

\paragraph{Scaling of downstream performance.}
~\autoref{fig:scaling}(c) reports the unweighted average zero-shot
raw accuracy over the nine task configurations in
~\autoref{sec:downstream-main}. Dual-Slot improves average accuracy
at every evaluated scale, with gains ranging from 0.58 to 2.27
percentage points for dense models and from 0.79 to 1.32 points
for MoE models. Together with the validation-loss results, these
findings demonstrate improvements in both language modeling and
downstream performance across the evaluated scales.

\subsection{Comparison with Parallel Loop Transformer}
\label{sec:plt-comparison}

We further compare Dual-Slot with Parallel Loop Transformer (PLT)
\citep{wu2025parallel}, another approach for parallelizing looped
computation. Following the dense scaling setup in~\autoref{sec:scaling},
we train PLT2 models at 150M, 300M, 600M, and 1B scales under the same
training settings and token budgets as the corresponding Vanilla and
Dual-Slot models.

\begin{figure}[!t]
    \centering
    \includegraphics[width=\linewidth]{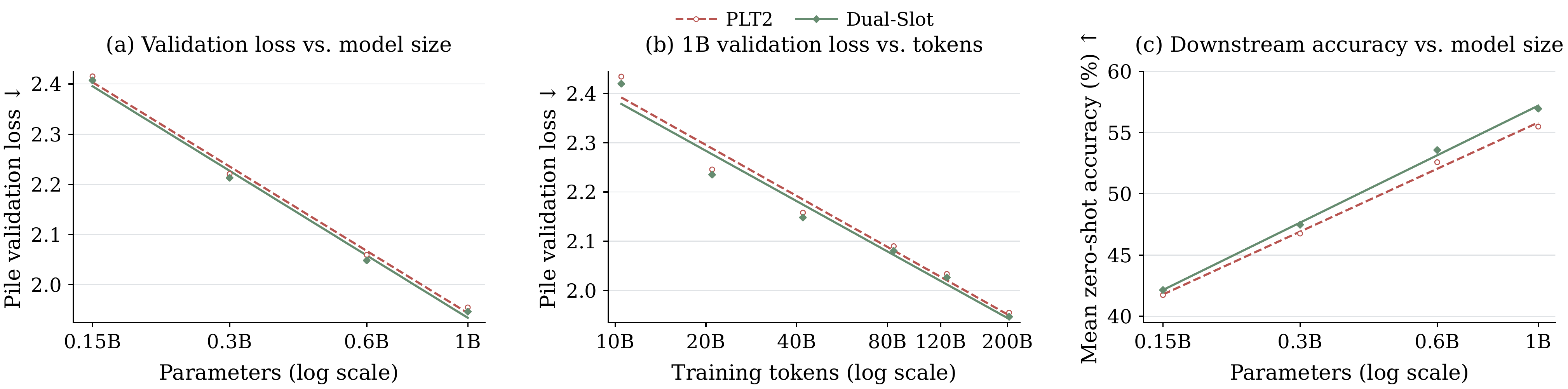}
    \caption{\textbf{Comparison with Parallel Loop Transformer (PLT2).}
    (a) Final Pile validation loss across dense model scales.
    (b) Pile validation loss over training tokens for the 1B models.
    (c) Average zero-shot raw accuracy over nine downstream task
    configurations across model scales.
    Dual-Slot consistently achieves lower validation loss and higher
    downstream accuracy than PLT2 across the evaluated scales.}
    \label{fig:plt-comparison}
\end{figure}

As shown in~\autoref{fig:plt-comparison}(a), Dual-Slot consistently
achieves lower Pile validation loss than PLT2 across all four model
sizes, with improvements of approximately 0.008--0.011.
\autoref{fig:plt-comparison}(b) further compares the training trajectories
of the 1B models, where Dual-Slot maintains lower validation loss
throughout training. These results indicate that the advantage of
Dual-Slot is not limited to the final checkpoint but persists over the
course of pretraining.

The improvement also transfers to downstream evaluation.
\autoref{fig:plt-comparison}(c) reports the unweighted average zero-shot
accuracy over the same nine task configurations used in
\autoref{sec:downstream-main}. Dual-Slot outperforms PLT2 at every
evaluated scale, with the gain increasing from 0.58 percentage points
at 150M to 1.61 points at 1B. Detailed results are reported in
\autoref{tab:plt-detailed}.

\begin{table}[!t]
    \centering
    \caption{\textbf{Detailed comparison with Parallel Loop Transformer.}
    Pile validation perplexity ($\downarrow$) and zero-shot raw accuracy
    (\%, $\uparrow$) across four dense model scales.
    Avg. denotes the unweighted mean over the nine downstream task
    configurations. Bold indicates the best result within each model scale.}
    \label{tab:plt-detailed}

    \scriptsize
    \setlength{\tabcolsep}{2.5pt}
    \renewcommand{\arraystretch}{1.13}

    \resizebox{\linewidth}{!}{
    \begin{tabular}{@{}lrrrrrrrrrrr@{}}
        \toprule
        \multirow{2}{*}{Method}
        & \multirow{2}{*}{\shortstack{Pile\\PPL}}
        & \multicolumn{2}{c}{LAMBADA}
        & \multirow{2}{*}{PIQA}
        & \multirow{2}{*}{\shortstack{Wino\\Grande}}
        & \multirow{2}{*}{SciQ}
        & \multirow{2}{*}{\shortstack{Hella\\Swag}}
        & \multicolumn{2}{c}{ARC}
        & \multirow{2}{*}{\shortstack{RACE\\high}}
        & \multirow{2}{*}{Avg.} \\
        \cmidrule(lr){3-4}
        \cmidrule(lr){9-10}

        &
        & OpenAI
        & Standard
        &
        &
        &
        &
        & Easy
        & Challenge
        &
        & \\
        \midrule

        \multicolumn{12}{c}{\textbf{150M}} \\
        \midrule

        Vanilla
        & 11.53
        & 37.34
        & 24.26
        & 62.24
        & 51.07
        & \textbf{75.80}
        & 29.32
        & 45.33
        & \textbf{19.37}
        & \textbf{29.38}
        & 41.57 \\

        PLT2
        & 11.19
        & \textbf{38.44}
        & 25.00
        & \textbf{64.25}
        & 50.20
        & 74.80
        & 29.28
        & 46.04
        & 19.11
        & 28.52
        & 41.74 \\

        Dual-Slot
        & \textbf{11.10}
        & 38.11
        & \textbf{27.05}
        & 63.93
        & \textbf{51.85}
        & 75.70
        & \textbf{29.89}
        & \textbf{46.63}
        & 18.77
        & 28.90
        & \textbf{42.32} \\

        \midrule

        \multicolumn{12}{c}{\textbf{300M}} \\
        \midrule

        Vanilla
        & 9.53
        & 45.62
        & 35.82
        & 66.38
        & 49.96
        & 78.40
        & 32.33
        & 49.16
        & 21.08
        & 30.24
        & 45.44 \\

        PLT2
        & 9.21
        & 48.61
        & 38.29
        & \textbf{66.70}
        & 51.78
        & 81.10
        & 33.32
        & 50.21
        & 20.56
        & 30.24
        & 46.76 \\

        Dual-Slot
        & \textbf{9.14}
        & \textbf{49.29}
        & \textbf{39.86}
        & 66.16
        & \textbf{51.93}
        & \textbf{81.50}
        & \textbf{33.33}
        & \textbf{51.18}
        & \textbf{21.33}
        & \textbf{31.58}
        & \textbf{47.35} \\

        \midrule

        \multicolumn{12}{c}{\textbf{600M}} \\
        \midrule

        Vanilla
        & 8.05
        & 56.53
        & 45.72
        & 69.10
        & 54.06
        & 86.00
        & 37.69
        & 56.65
        & 23.98
        & 32.06
        & 51.31 \\

        PLT2
        & 7.84
        & 58.47
        & 47.84
        & \textbf{69.97}
        & 53.67
        & 86.20
        & 38.82
        & \textbf{57.28}
        & \textbf{27.13}
        & 33.88
        & 52.58 \\

        Dual-Slot
        & \textbf{7.76}
        & \textbf{59.48}
        & \textbf{50.75}
        & 68.93
        & \textbf{57.22}
        & \textbf{88.60}
        & \textbf{39.15}
        & 57.03
        & 24.74
        & \textbf{34.64}
        & \textbf{53.39} \\

        \midrule

        \multicolumn{12}{c}{\textbf{1B}} \\
        \midrule

        Vanilla
        & 7.25
        & 62.64
        & 52.42
        & 71.93
        & \textbf{60.38}
        & 89.30
        & 42.09
        & 60.82
        & \textbf{27.56}
        & 35.31
        & 55.83 \\

        PLT2
        & 7.06
        & 63.30
        & 52.65
        & 72.25
        & 58.80
        & 87.80
        & 42.95
        & 60.94
        & 26.45
        & 34.26
        & 55.49 \\

        Dual-Slot
        & \textbf{7.00}
        & \textbf{64.95}
        & \textbf{56.69}
        & \textbf{72.58}
        & 59.75
        & \textbf{90.20}
        & \textbf{43.74}
        & \textbf{61.95}
        & 27.47
        & \textbf{36.56}
        & \textbf{57.10} \\

        \bottomrule
    \end{tabular}
    }
\end{table}

The per-task results in~\autoref{tab:plt-detailed} show that the average
improvement is not driven by a single benchmark. At 300M, 600M, and 1B,
Dual-Slot obtains the highest average downstream accuracy and improves
over PLT2 by 0.59, 0.81, and 1.61 percentage points, respectively.
At 1B, for example, Dual-Slot improves substantially over PLT2 on
LAMBADA Standard (56.69\% vs.\ 52.65\%), SciQ
(90.20\% vs.\ 87.80\%), HellaSwag (43.74\% vs.\ 42.95\%), and
RACE-high (36.56\% vs.\ 34.26\%). Although PLT2 remains stronger on a
small number of individual task configurations, Dual-Slot achieves the
best overall average at every evaluated scale.

\paragraph{Mathematical reasoning.}
We further compare the 1B models on GSM8K after supervised fine-tuning.
Under zero-shot evaluation, accuracy improves from 13.12\% for Vanilla
to 15.85\% for PLT2, while Dual-Slot reaches 18.95\%. Dual-Slot therefore
outperforms PLT2 by an additional 3.10 percentage points and Vanilla by
5.83 points. Under eight-shot prompting, PLT2 achieves 13.34\%, while
Dual-Slot further improves upon it by 1.60 percentage points. These
results suggest that the advantage of Dual-Slot extends beyond language
modeling perplexity and general zero-shot evaluation to supervised
mathematical reasoning.

Overall, Dual-Slot provides consistently stronger modeling quality than
PLT2 under matched model scales and training-token budgets, achieving
lower pretraining loss, higher average downstream accuracy, and better
GSM8K performance.

\subsection{Decoding Efficiency}
\label{sec:decoding-efficiency}

We evaluate the autoregressive decoding efficiency of Dual-Slot on a
single GPU using BF16, SDPA, and the standard KV cache.~\autoref{tab:decoding-efficiency} reports decoding throughput for the
1B dense and A0.6B MoE models across representative batch sizes.

Dual-Slot retains 93.1--96.5\% of Vanilla throughput for the 1B dense
model and 98.1--99.6\% for the A0.6B MoE model. These results indicate
that Dual-Slot introduces only modest decoding overhead. Throughput is measured over the decoding phase after
a 128-token prefill, with 1,024 generated tokens.

\begin{table}[!htbp]
    \centering
    \caption{\textbf{Autoregressive decoding throughput.}
    Throughput (tokens/s) on a single GPU using BF16, SDPA,
    and the standard KV cache. DS/V denotes the throughput ratio between
    Dual-Slot and Vanilla.}
    \label{tab:decoding-efficiency}

    \small
    \setlength{\tabcolsep}{5pt}
    \renewcommand{\arraystretch}{1.12}
    \begin{tabular}{@{}crrc@{\hspace{1.8em}}crrc@{}}
        \toprule
        \multicolumn{4}{c}{\textbf{1B Dense}}
        & \multicolumn{4}{c}{\textbf{A0.6B MoE}} \\
        \cmidrule(r){1-4}
        \cmidrule(l){5-8}

        BS
        & Vanilla
        & Dual-Slot
        & DS/V
        & BS
        & Vanilla
        & Dual-Slot
        & DS/V \\
        \midrule

        1
        & 80.51
        & 74.92
        & $0.9306\times$
        & 1
        & 17.25
        & 16.92
        & $0.9809\times$ \\

        4
        & 307.53
        & 292.29
        & $0.9504\times$
        & 4
        & 68.29
        & 67.97
        & $0.9953\times$ \\

        32
        & 2378.11
        & 2294.87
        & $0.9650\times$
        & 32
        & 549.12
        & 546.74
        & $0.9957\times$ \\

        128
        & 9527.89
        & 9056.98
        & $0.9506\times$
        & 128
        & 2191.15
        & 2162.76
        & $0.9870\times$ \\

        \bottomrule
    \end{tabular}
\end{table}

\subsection{Convergence of Jacobi Refinement}
\label{sec:convergence}

\begin{figure}[!t]
    \centering
    \includegraphics[width=\linewidth]{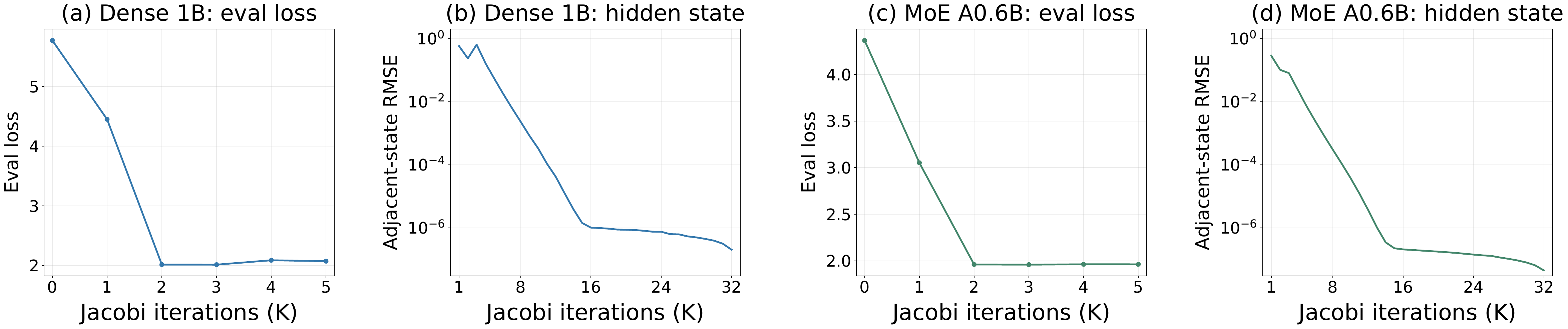}
    \caption{\textbf{Convergence of Jacobi refinement.}
    From left to right: evaluation loss and inter-iteration hidden-state
    RMSE for the 1B dense model, followed by the corresponding measurements
    for the A0.6B MoE model. Evaluation loss decreases rapidly up to the
    training depths ($K=2,3$) and changes comparatively little with further
    updates, while the RMSE between consecutive hidden states diminishes
    under continued refinement.}
    \label{fig:jacobi-convergence}
\end{figure}

We examine the iterative behavior of Dual-Slot through evaluation loss
and hidden-state changes as the number of Jacobi updates increases.~\autoref{fig:jacobi-convergence} reports both measurements for the
1B dense and A0.6B MoE models.

\paragraph{Evaluation loss.}
We evaluate both models with $K=0,\ldots,5$ Jacobi updates on 4096 samples. Evaluation
loss is high at $K=0$ and decreases rapidly as refinement proceeds,
reaching low values at $K=2$ and $K=3$, the depths used during training.
Further iterations introduce only small changes, with both models remaining near the loss levels reached
at their training depths.

\paragraph{Hidden-state convergence.}
We measure the change between consecutive iterations using the
root-mean-square error (RMSE):
\begin{equation}
    \mathrm{RMSE}^{(k)}
    = \frac{\left\|H^{(k)}-H^{(k-1)}\right\|_F}{\sqrt{N}},
    \label{eq:jacobi-state-error}
\end{equation}
where $H^{(k)}$ denotes the hidden states at iteration $k$ and $N$
is the number of scalar elements included in the comparison.
We continue increasing the number of Jacobi updates and observe that
the changes between consecutive hidden states become progressively
smaller for both models. At $K=32$, the RMSE reaches
$2.03\times10^{-7}$ for the dense model and $4.47\times10^{-8}$
for the MoE model, indicating empirical stabilization under
continued refinement.

Together, these observations provide empirical evidence that training Dual-Slot with
only two or three Jacobi updates is sufficient to learn stable and
convergent iterative dynamics for models.

\section{Ablation Studies}
\label{sec:ablation}

We conduct all ablations on the 600M dense model trained for 120B tokens.
We examine three design choices: how to construct the prediction-slot
input, where to place the loop core, and whether to use fixed or random
$K$ during training.

\subsection{Prediction-Slot Fusion}
\label{sec:fusion-ablation}

We compare four prediction-slot fusion methods: direct addition (Raw Add),
postmerge RMS matching, gated addition, and gated addition with postmerge
RMS matching. RMS matching rescales the merged representation to match
the RMS of $x_t$, while the learnable gate controls the contribution of
$h_{t-1}$ before addition.

As shown in~\autoref{tab:fusion-ablation}, RMS matching increases
validation loss both with (w/)  and without (w/o)  gating. Gated addition improves over Raw Add by only 0.001. We therefore
retain Raw Add as our default for its simplicity and parameter-free
formulation.

\begin{figure}[!t]
    \centering
    \includegraphics[width=\linewidth]
        {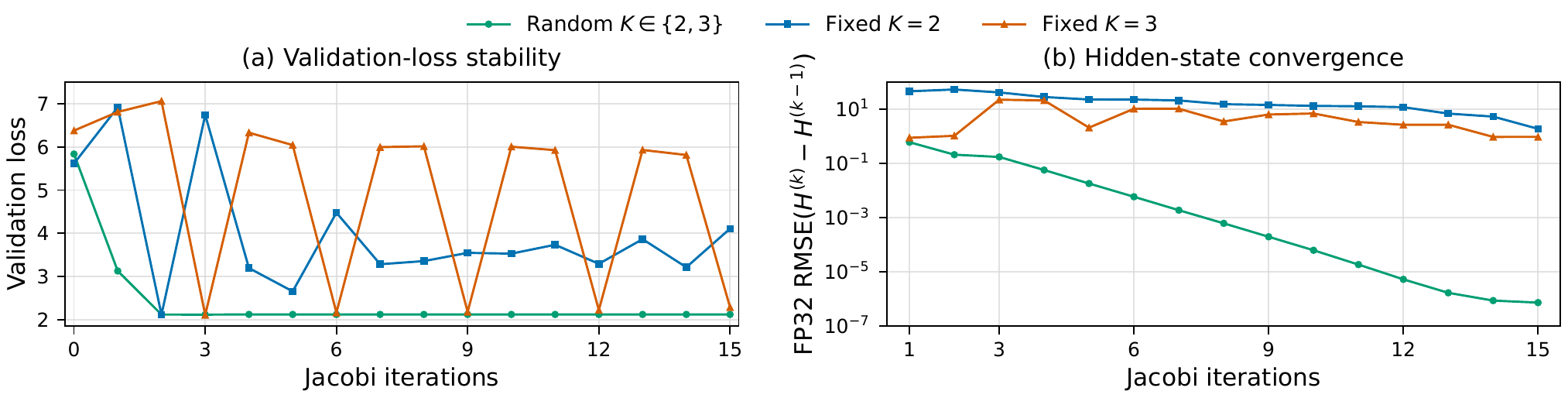}
    \caption{\textbf{Fixed versus random Jacobi iterations.}
    Validation loss (left) and inter-iteration hidden-state RMSE (right)
    for models trained with fixed $K=2$, fixed $K=3$, or random
    $K\in\{2,3\}$. With random $K$, loss changes only slightly beyond
    the training iteration counts, and hidden states stabilize faster
    within the evaluated range.}
    \label{fig:jacobi-depth-ablation}
\end{figure}

\subsection{Loop-Core Location}
\label{sec:core-location}

With the loop-core depth fixed to one third of the model, we investigate
whether the core should be placed in the first, middle, or final third.

As shown in~\autoref{tab:core-location}, all three placements improve
over Vanilla, but the middle-third configuration achieves the lowest
validation loss. It reduces loss from 2.0859 to 2.0485, outperforming
the first- and final-third configurations by 0.0105 and 0.0328,
respectively. These results support our choice of the middle third
as the default loop-core location.

\begin{table}[t]
    \centering
    \small
    \renewcommand{\arraystretch}{1.12}

    \sbox{\fusioncaptionbox}{%
        \begin{minipage}[t]{0.48\textwidth}
            \caption{\textbf{Prediction-slot fusion.}
                Validation loss with (w/) and without (w/o) learnable gating on the
                600M dense model trained for 120B tokens. Loss gaps are
                measured relative to Raw Add without a gate; negative
                values indicate improvement. The best loss in each
                group is bold.}
            \label{tab:fusion-ablation}
        \end{minipage}%
    }
    \sbox{\locationcaptionbox}{%
        \begin{minipage}[t]{0.48\textwidth}
            \caption{\textbf{Loop-core location.}
    Validation loss for different placements of a seven-layer Dual-Slot
    core in a 21-layer Transformer.
    Loss gaps are measured relative to Vanilla; negative values indicate
    improvement. The best validation loss is bold.}
            \label{tab:core-location}
        \end{minipage}%
    }

    \setlength{\sharedcaptionheight}{%
        \dimexpr\ht\fusioncaptionbox+\dp\fusioncaptionbox\relax
    }
    \ifdim
        \dimexpr\ht\locationcaptionbox+\dp\locationcaptionbox\relax
        >\sharedcaptionheight
        \setlength{\sharedcaptionheight}{%
            \dimexpr\ht\locationcaptionbox+\dp\locationcaptionbox\relax
        }
    \fi

    \begin{minipage}[t]{0.49\textwidth}
        \centering
        \parbox[t][\sharedcaptionheight][t]{\linewidth}{%
            \usebox{\fusioncaptionbox}%
        }\par
        \begin{tabular*}{\linewidth}{
            @{\extracolsep{\fill}}l@{}r@{}r@{}
        }
            \toprule
            Method & Val. loss $\downarrow$ & Loss gap \\
            \midrule
            Raw Add w/o gate
            & \textbf{2.0485} & $0.0000$ \\
            RMS matching w/o gate
            & 2.0497 & $+0.0012$ \\
            Raw Add w/ gate
            & \textbf{2.0474} & $-0.0011$ \\
            RMS matching w/ gate
            & 2.0490 & $+0.0006$ \\
            \bottomrule
        \end{tabular*}
    \end{minipage}
    \hfill
    \begin{minipage}[t]{0.49\textwidth}
        \centering
        \parbox[t][\sharedcaptionheight][t]{\linewidth}{%
            \usebox{\locationcaptionbox}%
        }\par
        \begin{tabular*}{\linewidth}{
            @{\extracolsep{\fill}}l@{}r@{}r@{}
        }
            \toprule
            Method & Val. loss $\downarrow$ & $\Delta$ vs. Vanilla \\
            \midrule
            Vanilla
            & 2.0859 & $0.0000$ \\
            Front7
            & 2.0590 & $-0.0269$ \\
            Middle7 (default)
            & \textbf{2.0485} & $\mathbf{-0.0374}$ \\
            Last7
            & 2.0813 & $-0.0046$ \\
            \bottomrule
        \end{tabular*}
    \end{minipage}
\end{table}

\subsection{Fixed versus Random Jacobi Iterations}
\label{sec:jacobi-depth-ablation}

We compare training with fixed $K=2$, fixed $K=3$, and random
$K\in\{2,3\}$ to examine the effect of randomizing the iteration count.
Following~\autoref{sec:convergence}, we evaluate each trained model
with increasing numbers of Jacobi iterations and examine both
validation loss and inter-iteration hidden-state changes. As shown in~\autoref{fig:jacobi-depth-ablation}, models trained with
a fixed $K$ are more sensitive to the number of Jacobi iterations used
at evaluation time. In contrast, the model trained with random $K\in\{2,3\}$
shows only minor changes in validation loss as the number of iterations
increases beyond those used during training.

The hidden-state results show the same trend. As additional Jacobi
iterations are applied, the model trained with random $K$ exhibits
substantially faster decay in the change between consecutive hidden
states, reaching an inter-iteration RMSE of $7.36\times10^{-7}$ at
$K=15$, compared with 1.90 and 0.96 for fixed $K=2$ and fixed $K=3$,
respectively.

Together, these results suggest that randomizing $K$ during training
reduces dependence on a particular iteration count and encourages
more stable iterative dynamics. We therefore use random $K\in\{2,3\}$
as the default training strategy.

\section{Conclusion}

We introduced Dual-Slot, an architecture that reorganizes serial looped
computation into parallel latent and prediction slots. By carrying latent
states across adjacent tokens, the two slots advance through shared layers
in a single paired invocation while preserving standard autoregressive
generation. A small number of Jacobi iterations preserves token-level
parallelism during training, while randomizing $K\in\{2,3\}$ encourages
more stable predictions and hidden-state refinement beyond the iteration
counts used during training. The experiment results highlight cross-token latent-state reuse
as a practical way to improve language model quality while avoiding
the within-token serial loop dependency of conventional looped models.

\section*{Acknowledgement}
This work is sponsored by the National Natural Science Foundation of China (NSFC) grant (No. 62576211) and the Specialized Program on Fundamental Research from Science and Technology Commission of Shanghai Municipality (No. 2025SHZDZX025G09).


\clearpage
\bibliography{references}
\bibliographystyle{plainnat}

\clearpage
\appendix

\section{Model Configurations and Training Details}
\label{app:training_details}

\subsection{Model Architecture}

\paragraph{Common backbone.}
All models are trained from random initialization and use a Qwen3-style decoder-only Transformer. The backbone combines grouped-query attention (GQA) \citep{ainslie2023gqa}, Q/K normalization, pre-RMSNorm \citep{zhang2019root}, SwiGLU with SiLU activations \citep{shazeer2020glu}, and rotary position embeddings (RoPE) \citep{su2024roformer}. Attention projections have no bias terms. The vocabulary contains 50,304 tokens, and the input embedding and output language-model head share their weights. Unless stated otherwise, we set \texttt{rms\_norm\_eps}$=10^{-6}$, \texttt{rope\_theta}$=10^{6}$, \texttt{attention\_dropout}$=0$, and \texttt{initializer\_range}$=0.02$. The attention head dimension is 64 at every model size.

At each scale, Vanilla and Dual-Slot have identical parameter counts and physical Transformer layers. Dual-Slot introduces no additional parameters. Layer indices in this paper start at zero, whereas the implementation indexes the core’s starting layer from one.

\subsubsection{Dense Models}

\autoref{tab:dense_configs} lists the dense configurations. Depth, hidden dimension, and FFN width increase with model size, while the Dual-Slot core occupies the middle third of the layers.

\begin{table}[!htbp]
\centering
\small
\caption{Dense model configurations. Vanilla and Dual-Slot have identical physical architectures and parameter counts. Dual-Slot core layers use zero-based indices.}
\label{tab:dense_configs}
\resizebox{\textwidth}{!}{
\begin{tabular}{@{}ccccccccc@{}}
\toprule
Model &
Parameters &
Layers &
$d_{\mathrm{model}}$ &
$d_{\mathrm{ff}}$ &
Q heads &
KV heads &
Dual-Slot core &
Prefix/Core/Suffix \\
\midrule
150M & 149,541,120 & 12 & 768 & 3,328 & 12 & 4 & L4--L7 & 4/4/4 \\
300M & 303,203,200 & 15 & 1,024 & 4,608 & 16 & 4 & L5--L9 & 5/5/5 \\
600M & 604,627,328 & 21 & 1,280 & 5,632 & 20 & 5 & L7--L13 & 7/7/7 \\
1B & 1,011,626,496 & 24 & 1,536 & 7,168 & 24 & 6 & L8--L15 & 8/8/8 \\
\bottomrule
\end{tabular}
}
\end{table}

\subsubsection{Mixture-of-Experts Models}

As shown in~\autoref{tab:moe_configs}, MoE models are based on the dense configurations. Each Transformer layer has 64 routed experts, selects eight experts for each token, and has no shared expert. Each expert has 1/8 of the corresponding dense model’s FFN width. Thus, the eight selected experts together have the same FFN width as the dense model. The two models have similar active parameter counts; the small difference mainly comes from the router.
The MoE models use the same attention, normalization, positional encoding, vocabulary, tied embeddings, and Dual-Slot core placement as the dense models. Their output projections use Qwen-style unscaled initialization.

\begin{table}[t]
\centering
\small
\caption{MoE model configurations. Active parameters follow the conventional count for one slot with Top-8 routing. Each model has 64 routed experts, and core layers use zero-based indices.}
\label{tab:moe_configs}
\resizebox{\textwidth}{!}{
\begin{tabular}{@{}ccccccccc@{}}
\toprule
Model &
Active params. &
Total params. &
Layers &
$d_{\mathrm{model}}$ &
$d_{\mathrm{ff}}^{\mathrm{expert}}$ &
Q/KV heads &
Dual-Slot core &
Prefix/Core/Suffix \\
\midrule
A0.15B & 150,130,944 & 794,218,752 & 12 & 768 & 416 & 12/4 & L4--L7 & 4/4/4 \\
A0.3B & 304,186,240 & 1,790,542,720 & 15 & 1,024 & 576 & 16/4 & L5--L9 & 5/5/5 \\
A0.6B & 606,347,648 & 3,785,499,008 & 21 & 1,280 & 704 & 20/5 & L7--L13 & 7/7/7 \\
\bottomrule
\end{tabular}
}
\end{table}

\subsection{Training Setup}
\label{app:training_setup}

All models are pretrained with the causal next-token prediction objective on the Pile dataset. Training sequences contain 2,048 tokens. The global batch comprises 1,024 sequences, or 2,097,152 tokens per optimizer step, and the per-device micro-batch contains eight sequences.

Dense models use AdamW \citep{loshchilov2017decoupled}. MoE models use Megatron Adam with the distributed optimizer \citep{narayanan2021efficient}. Both use $\beta_1=0.9$, $\beta_2=0.95$, $\epsilon=10^{-8}$, weight decay of $0.1$, and global gradient-norm clipping at $1.0$. The learning rate warms up linearly over the first 5\% of training, then follows a cosine schedule to 10\% of its peak value. The peak is $1.5\times10^{-3}$ for the 150M, 300M, and 600M dense models and their MoE counterparts; $1.0\times10^{-3}$ for the 1B dense. For MoE models, the Megatron optimizer maintains the main parameters and accumulated gradients in FP32. All runs use random seed 42, with activation recomputation disabled. At a given scale, Vanilla and Dual-Slot share the physical architecture, batch size, optimizer hyperparameters, learning-rate schedule, and training protocol unless an ablation specifies otherwise.

\section{Supervised Fine-Tuning Details}
\label{app:sft_details}

We perform full-parameter supervised fine-tuning (SFT) for mathematical
reasoning on GSM8K. Vanilla and Dual-Slot start from their respective
pretrained checkpoints and use identical fine-tuning data and
hyperparameters. Dual-Slot retains its pretraining recurrence, with
the number of Jacobi updates $K$ sampled from $\{2,3\}$ for each
training batch.

GSM8K SFT uses three epochs, a global batch size of 128, a maximum
sequence length of 2,048, and no sequence packing. Optimization uses
AdamW with a peak learning rate of $2\times10^{-5}$, $\beta_1=0.9$,
$\beta_2=0.95$, $\epsilon=10^{-8}$, zero weight decay, and global
gradient-norm clipping at $1.0$. The learning rate follows a cosine
schedule with a warmup ratio of $0.03$. Training uses BF16 and random
seed 42. Only response tokens contribute to the loss; prompt tokens
are masked.

We use the official GSM8K training split of 7,473 examples. The
official test split of 1,319 examples is used exclusively for
evaluation. Each training input places the problem after the label ``Question:''
and starts the next line with ``Answer:''; the original GSM8K
response serves as the target. The target includes the
intermediate reasoning trace and the final answer marked by
\texttt{\#\#\#\#}.

\end{document}